%% file: main.tex
\documentclass[nohyperref]{article}
\usepackage{graphicx}
\usepackage[accepted]{icml2022}
\usepackage{algorithmic, algorithm}
\usepackage{natbib}

\usepackage{pbox}

\begin{document}

%

%

\icmltitlerunning{Learning to Simulate Unseen Physical Systems}

\twocolumn[

\icmltitle{Learning to Simulate Unseen Physical Systems \\ with Graph Neural Networks}
\begin{icmlauthorlist}
    \icmlauthor{Ce Yang}{Tsinghua}
    \icmlauthor{Weihao Gao}{ByteDance}
    \icmlauthor{Di Wu}{ByteDance}
    \icmlauthor{Chong Wang}{ByteDance_US}
\end{icmlauthorlist}

\icmlaffiliation{Tsinghua}{Tsinghua University, Beijing, China}
\icmlaffiliation{ByteDance}{ByteDance Inc., Beijing, China}
\icmlaffiliation{ByteDance_US}{ByteDance Inc., Bellevue, WA, USA}
\icmlcorrespondingauthor{Weihao Gao}{weihao.gao@bytedance.com}
\vskip 0.3in
]

\printAffiliationsAndNotice{} 

\begin{abstract}
Simulation of the dynamics of physical systems is essential to the development of both science and engineering. Recently there is an increasing interest in learning to simulate the dynamics of physical systems using neural networks. However, existing approaches fail to generalize to physical substances not in the training set, such as liquids with different viscosities or elastomers with different elasticities. Here we present a machine learning method embedded with physical priors and material parameters, which we term as “Graph-based Physics Engine” (GPE), to efficiently model the physical dynamics of different substances in a wide variety of scenarios. We demonstrate that GPE can generalize to materials with different properties not seen in the training set and perform well from single-step predictions to multi-step roll-out simulations. In addition, introducing the law of momentum conservation in the model significantly improves the efficiency and stability of learning, allowing convergence to better models with fewer training steps.
\end{abstract}

\section{Introduction}

\input{introduction}

\subsection{Related Work}
\label{sec:related_work}
\input{related_work}

\section{Model Framework}
\label{sec:model}
\input{model_framework}

\section{Experiment Results}
\label{sec:exp}
\input{experiments}



\section{Conclusion}
\label{sec:conclusion}
\input{conclusion}


\bibliography{reference.bib}

\end{document}

%% file: introduction.tex

Simulation of the dynamic of unknown physics systems is crucial in many scientific and engineering areas~\cite{belytschko1994element, SULSKY1995236, jiang2016material,sifakis2012fem, han2017deep, holl2020learning}. Building high precision physics simulators requires extensive domain knowledge and substantial engineering effort. However, the approximation techniques used to ensure perceptual realism make the simulations deviate from the laws of reality in the long term. 
Recently, several neural network based simulators were proposed to learn physical dynamics from large-scale training data~\cite{battaglia2016interaction,chang2016compositional,li2018learning,mrowca2018flexible,sanchez2018graph,li2019propagation,ummenhofer2019lagrangian,sanchez2020learning}. Their successes prove the feasibility of using machine learning to improve physics simulation engines.

One common drawback of the aforementioned methods is the lack of generalization for unseen physical processes and substances. For example, if the training set consists of dynamics of a softer elastomer and a harder one, it is expected that the model can generalize to elastomers with other levels of elasticity. However, most machine learning based simulators fail to do so since they can only simulate the physic systems already seen in the training set. Also, machine learning based simulators often suffer from instability in long-range roll-out simulation and inefficiency in training. Hence, physical simulators that are generalizable to unseen physical environments, stable in long-range simulation and efficient to train are highly desired for practical applications.


To this end, we introduce a novel machine learning approach to model physical dynamics with graphs, which we term as “Graph-based Physics Engine” (GPE). GPE uses a graph structure to represent discrete physical systems where the force interactions are modeled using a message passing neural network (MPNN)~\cite{gilmer2017neural}. The contribution of GPE is multifaceted. First, the physical parameters such as elasticity of elastomers are introduced to enable the model to generalize across systems with different parameters. Second, a unified model architecture with different graph construction is used to learn different types of physical dynamics. This enables GPE to generalize to a wide variety of systems from viscous fluid simulations and deformation dynamics of discrete substances to elastomeric deformation and finite element mechanics (FEM) analysis. Additionally, in order to improve stability of roll-out simulation and efficiency of training, we introduce an important or even indispensable law --- the law of momentum conservation, by modifying the message passing process in GPE. 

Various experiment results show that, by introducing the physical parameters, GPE can generalize well to physical systems and materials not seen in the training set, as well as larger systems with more complex boundaries, without any adjustments. We also demonstrate that the introduction of momentum conservation significantly improves training stability and computational efficiency. Additionally, GPE shows good long-term roll-out stability even if it is learned over single-step loss. We believe that GPE can provide new insights into the construction of machine learning based simulators, and is a key step toward predicting unknown physics problems in the real world. 

The rest of the paper is organized as follows. 
In Section~\ref{sec:model}, we introduce the model framework of GPE, and show how the generalization of physical parameters, adaptation to different physical systems and momentum conservation are incorporated in GPE. In Section~\ref{sec:exp}, we demonstrate the performance of GPE by visualizing the predicted dynamics of various domains. We also compare the results with several baseline models and study the effect of momentum conservation. We conclude our paper and discuss some future directions and societal impacts in Section~\ref{sec:conclusion}.

%% file: related_work.tex
Learning dynamics of physical systems from data is one of the most important research areas in physics \cite{10.1145/280814.280816}. A variety of neural network architectures, such as recurrent neural networks~\cite{kadupitiya2020deep}, convolutional neural networks~\cite{ummenhofer2019lagrangian} and graph neural networks~\cite{sanchez2018graph, battaglia2016interaction, li2018learning, li2019propagation}, have been adopted to simulate the dynamics of physical systems. In particular, Ummenhofer et al.~\cite{ummenhofer2019lagrangian} developed a novel CNN model based on smoothed-particle hydrodynamics (SPH) theory to simulate the motion of a Lagrangian fluid.
Battaglia et al.~\cite{battaglia2016interaction, battaglia2018relational} and Sanchez-Gonzalez et al.~\cite{sanchez2020learning, sanchez2018graph, sanchez2019hamiltonian} demonstrated the effectiveness and generality to model the dynamics by message passing. In these works, similar network architectures were used for different physic systems including multi-body systems, robot control, and deformation simulations for fluid and rigid bodies. 
On the other hand, Mrowca et al.~\cite{mrowca2018flexible} and Li et al.~\cite{li2018learning, li2019propagation}  used hierarchical graph neural networks to simulate non-rigid deformation and collision deformation of structured physical systems such as elastomers. However, all the aforementioned work can only simulate the systems seen in the training set, which motivates us to allow the learned model to simulate unseen systems.


%% file: model_framework.tex


Assume $(G^0, G^1, ..., G^T)$ is the evolutionary trajectory of a physical system over time, where a directed graph $G^t=<O^t, R^t> \in \mathcal{G}$ represents the state of the system at time $t$. We may omit the superscript $t$ whenever it does not introduce ambiguity.
Vertices $O = \{o_i^{\xi}\}$ represent the collection of discrete units of objects, where each node $o_i^{\xi}$ represents a single unit
and the superscript $\xi$ indicates the material type of the unit. Edges $R=\{(o_i^{\xi}, o_j^{\eta})\}$ represent the (directed) interaction between the units, i.e.,  $(o_i^{\xi}, o_j^{\eta}) \in R$ means that there exists an interaction between node $o_i^{\xi}$ and node $o_j^{\eta}$.

A learnable graph-based physics engine (GPE) $s_\theta:\mathcal{G} \rightarrow \mathcal{G}$ models the dynamics of a physical system by mapping the graph representation $G^t$ to the graph representation at the next time step $G^{t+1}$. Similar to~\cite{sanchez2020learning}, we use message passing neural networks (MPNN) to model the evolutionary dynamics of physical systems. The details of the algorithm in shown in Algorithm~\ref{alg:mpnn}.

\begin{algorithm}
   \caption{MPNN used in graph-based physics engine (GPE).}
   \label{alg:mpnn}
\begin{algorithmic}
   \STATE \textbf{Input:} raw features of nodes $o_i^{\xi}$ and edges $(o_i^{\xi}, o_j^{\eta})$.
   \STATE \textbf{Encoder}:
   \STATE $h_i^{\xi,0} = f_{v,0}^{\xi}({\rm features}(o_i^{\xi}))$ for all $o_i^{\xi} \in O$.
   \STATE $h_{i, j}^0 = f_{e,0}^{(\xi,\eta)}({\rm features}(o_i^{\xi}, o_j^{\eta}))$ for all $(o_i^{\xi}, o_j^{\eta}) \in R$.
   \STATE \textbf{Processor:}
   \FOR{$l = 1, \dots , L$}
      \STATE $m_{i, j}^l = f_{e, l}^{(\xi, \eta)}(h_i^{\xi, l-1}, h_j^{\eta, l-1}, h_{i, j}^{l-1})$ for all $(o_i^{\xi}, o_j^{\eta}) \in R$.
      \STATE $m_{j, i}^l = -m_{i, j}^l$ for all $(o_i^{\xi}, o_j^{\eta}) \in R$. 
      \STATE $h_i^{\xi, l} = h_i^{\xi, l-1} + f_{v, l}^{\xi} (\sum_{j \in \mathcal{N}_i} m_{i,j}^l)$ for all $o_i^{\xi} \in O$.
      \STATE $h_{ij}^l = h_{i, j}^{l-1} + m_{i, j}^l$ for all $(o_i^{\xi}, o_j^{\eta}) \in R$. 
    \ENDFOR
    \STATE \textbf{Decoder:}
    \STATE $a_i^{\xi} = f_v^{\xi} (h_i^{\xi,L})$ for all $o_i^{\xi} \in O$.
\end{algorithmic}
\end{algorithm}

In the \textbf{encoder}, the raw features of the nodes $o_i^{\xi}$ and the edges $(o_i^{\xi}, o_j^{\eta})$ are mapped into hidden vectors $h_i^{\xi,0}$ and $h_{i, j}^0$. GPE defines each material as a specific node type in the MPNN, which posed a strong inductive bias that nodes of the same type should have the same dynamics and vice versa. Then, the \textbf{processor} iteratively calculates the delivered messages $m_{i, j}^l$ on the edges based on hidden vectors of the source node, the destination node and the edge itself. The backward message is just the negative of the forward message. Then, the node hidden vectors $h_i^{\xi, l}$ are update by aggregating the edge messages at the previous step, and the edge hidden vectors $h_{ij}^l$ are update by adding the edge message. Finally, the \textbf{decoder} uses the hidden vectors of each node to predict the position at the next time step. Note that the processor and decoders used for different types of nodes and edges are also different. 


\subsection{Input Feature}

The first contribution of GPE is to introduce physical parameters in the input feature. The input features on GPE nodes contain the velocity of the unit, and the external forces applied such as gravity, and the material properties of each unit (such as the viscosity of fluid, the friction angle of sand, the hardening coefficient of snow, the Young's modulus of an elastomer). Unlike the work of Sanchez-Gonzalez et.al.~\cite{sanchez2018graph} which learns distinct embeddings for different materials, we adopt a \textit{continuous interpolation} of materials embeddings. Take the viscous fluid system as an example. If the embedding for fluid unit with viscosity $v_1$ and $v_2$ are $\Theta(v_1)$ and $\Theta(v_2)$ respectively, then the embedding for fluid unit with other levels of viscosity is given by
$$\Theta(\lambda v_1 + (1-\lambda) v_2) = \lambda \Theta(v_1) + (1-\lambda) \Theta(v_2).$$
Thanks to the ability of interpolation of neural networks, GPE enjoys better generalization performance for scenarios and materials not seen in the training set, which will be shown later. Position information is not included in the input feature, which makes GPE naturally satisfy the translation invariance. The edge feature contains only the relative distance of the two endpoints, which maintains permutation invariance.

\subsection{Graph Representation of Physical Systems}

In order to use a unified model architecture to learn the dynamics of different types of physical systems, we adopt different graph topology depending on how the physical system is being discretized. Two common approaches to discretize physical systems are \textbf{particle-based} discretization and \textbf{mesh-based} discretization. Particle-based discretization is widely used in physical systems without fixed shape or connectivity, such as liquids, sand and snow. With better representation ability but more restrictions on the deformation evolution, mesh-based discretization are usually applied to physical systems with connectivity such as solids and elastomers in finite element analysis.

\begin{figure}
    \centering
    \includegraphics[width=0.48\textwidth]{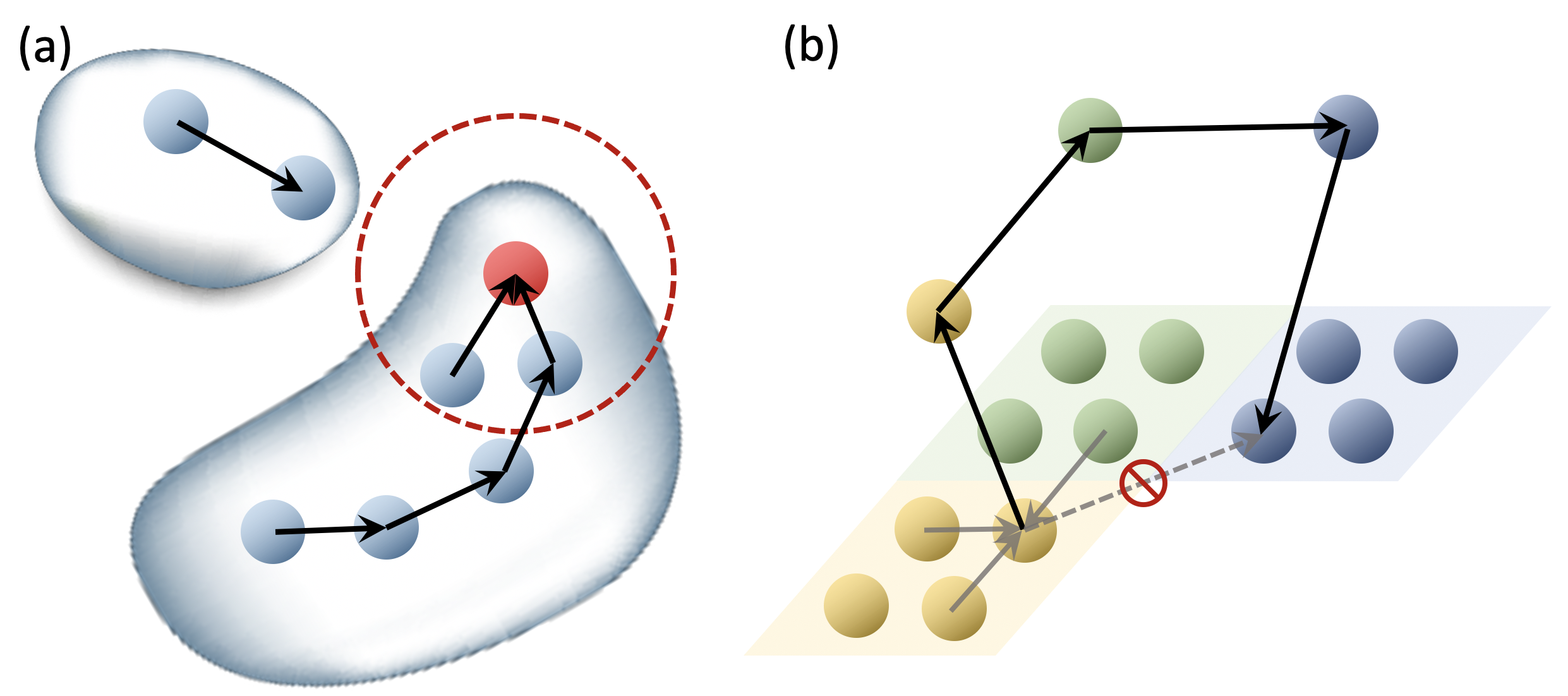}
    \caption{Graph topology for particle-based and mesh-based discretized physical systems. (a) Dynamic nearest neighbor graph for particle-based discretized systems. Two nodes are connected if their distance is less than some threshold. (b) Multi-scale grid graph for mesh-based descritized systems. Nodes in the same mesh cells form a virtual node and are connected by virtual edges.}
    \label{fig:graph_topo}
\end{figure}

In GPE, \textbf{particle-based} discretized systems are modeled by dynamic nearest neighbor graphs where two particles $o_i^{\xi}$ and $o_j^{\eta}$ are connected if their distance is less than a threshold $r$. The nearest neighbor topology reflects the fact that the strength of interaction between two particles vanishes as their distance increases. Due to the deformation of the system during roll-out simulation, the nearest neighbor graph needs to be updated for each time step.

In contrast, \textbf{mesh-based} discretized systems are modeled by a static multi-scale grid graph that is consistent across all time steps. In such a graph, one node is set for each mesh point, and only nodes in the same mesh cell are connected by edges to reflect the structure of the mesh. In order to take consideration of the monolithicity and long-range interaction during the deformation of solids, we use a multi-scale grid graph topology similar to~\cite{mrowca2018flexible,li2018learning}. Taking a two-dimensional structure as an example, we merge four meshes in the same mesh cell into one ``macromesh'' and use a new ``virtual node'' to represent the macromesh, and define the connectivity between virtual nodes as ``virtual edges'' in the graph. Furthermore, four neighboring macromeshes can be further merged into a larger macromesh, until the whole system is merged into one largest macromesh. These virtual nodes and virtual edges only serve to describe the macro structure and build a high-speed path for force transmission, rather than representing real meshes.

\subsection{Momentum Conservation}

An effective method to improve the stability of machine learning models is to incorporate the fundamental laws of the underlying problem. For example, incorporating \textit{translation invariance (equivariance)} for convolutional neural networks~\cite{bruna2013spectral,defferrard2016convolutional} and \textit{permutation invariance (equivariance)} for graph neural networks~\cite{kipf2018neural,satorras2021n} have been proved to successfully improve stability of the models. The most fundamental law of dynamic systems is the law of \textbf{momentum conservation}, which is a direct corollary of Newton's law of motion. However, this law is omitted by many previously developed machine learning based physical simulators. Hereby, we adjust the message passing method of MPNN to ensure the conservation of momentum, and achieve significant improvement in computational efficiency and training stability.


To ensure conservation of momentum, we incorporate Newton's third law of motion which states that all forces between two objects exist in equal magnitude and opposite direction. In message passing neural network, we need to make sure that the message passed on the edge $e_{i \rightarrow j}$ equals to negative of the message passed on the edge $e_{j \rightarrow i}$, at any time $t$. Notice that either the nearest neighbor graph or multi-scale grid graph contains bi-directional edges between connected nodes, i.e., if $e_{i \rightarrow j} = (o_i^{\xi}, o_j^{\eta}) \in R$, then $e_{j \rightarrow i} = (o_j^{\eta}, o_i^{\xi}) \in R$. So we can simply let $\vec m_{j \rightarrow i}^t = -\vec m_{i \rightarrow j}^t$ rather than computing each message on a directed edge separately. This simple adjustment not only ensures momentum conservation but also reduces the computation of messages in the graph convolutional network by a half without changing the number of parameters of the model, which accounts for the majority of the computation. We will show that GPE enjoys better stability with the introduction of momentum conservation.

%% file: experiments.tex
In this section, we demonstrate the performance of GPE in a wide variety of physical systems, including both particle-based and mesh-based discretized systems. We will show that GPE can generalize to unseen physical systems and show the improvement of introducing momentum conservation.

\subsection{Physical Domains and Datasets}
\textbf{Particle-based} systems include several physical domains with different behaviors: \textsc{Fluid}, \textsc{Sand} and \textsc{Snow}. \textsc{Fluid} domain contains dynamics of viscous fluids which is an approximate chaotic system whose internal damping is highly correlated with the viscosity. The \textsc{sand} domain contains dynamics of incompressible granular materials with different frictional behavior, depending mainly on its friction angle. The \textsc{snow} domain contains the deformation of snow which can be modeled by a compact solid by volume compression. The tensile deformation may cause the snow to fracture, which depends mainly on its hardening coefficient. We will test the generalizability of GPE by training the model on small and simple 2D systems and testing the model on large 2D or 3D systems with complex boundaries.

\textbf{Mesh-based} systems include the movement and deformation of 
elastomers, which depends not only on their material properties, but also on their structural shape to ensure structural integrity at local levels. We constructed the \textsc{elastomer} domain and the \textsc{FEM} domain for different behaviors of elastomers respectively. The \textsc{elastomer} domain contains dynamics of elastomers with different elasticities hitting on a flat surface. The \textsc{FEM} domain contains the finite element analysis of the collision behavior of elastomers on irregular obstacles, as well as the internal stress distribution. It is important to note that for both particle-based systems and mesh-based systems, each domain contains data for materials with a range of different material parameters (the viscosity of fluid, the friction angle of sand, the hardening coefficient of snow, the Young's modulus of the elastomer).

\textbf{Dataset generation}. The datasets for \textsc{Sand}, \textsc{Snow} and \textsc{Elastomer} domains are generated by the Taichi-MPM engine~\cite{hu2019taichi,hu2019difftaichi}. The dataset for \textsc{Fluid} domain is generated by the FLIPViscosity3D engine~\cite{batty2007fast,batty2008accurate}. For each domain, we independently sample 2000 trajectories with 10 different physical parameters as training set, 300 trajectories as test set and 300 trajectories with unseen physical parameters as validation set, where each trajectory contains 1000 frames. The \textsc{Fluid} domain contains about 5000 particles and the other three domains contain approximately 2500 particles. The \textsc{FEM} domain is obtained from standard simulation formulas. We sample 2000 trajectories as training set, 300 trajectories as test set and 300 trajectories as validation set, where each trajectories contain 800 frames. The \textsc{FEM} domain approximately contains 1000 meshes. The model is trained on data with Young's modulus from 1 MPa to 4 MPa, and tested on data with Young's modulus from 0.5 MPa to 10 MPa.

 \subsection{Implementation Details}
\textbf{Details of model architecture}. The components of GPE are implemented by multi-layer perceptrons (MLP). The encoders $\{f_{v,0}^{\xi}, f_{e,0}^{(\xi, \eta)}\}$, the processors $\{f_{v,l}^{\xi}, f_{e,l}^{(\xi, \eta)}\}$ and the decoders $f_{v}^{\xi}$ for every node type are all MLPs with 2 layers and 128 neurons per layer. The typical value of the number of message passing steps is 12. All activation functions are PReLU~\cite{he2015delving}. For \textsc{Sand}, \textsc{Snow} and \textsc{Fluid} domains, the cutoff radius for the nearest neighbor graph is 0.4.

\textbf{Training details}. 
In roll-out simulations, the accumulation of errors will lead to significant deviations from the ground truth in the long-term. To accommodate this phenomenon, we added the same Gaussian noise to the input position and the target position of each particle in training set. The standard deviation of the Gaussian noise ranges from $5\times 10^{-5}$ to $1\times 10^{-2}$ for different domains. We use one-step training, where the model is trained by the mean square error (MSE) of the predicted positions and the ground truth at the next time step.  The model is trained 1000 epochs, using the Adam optimizer~\cite{kingma2014adam} at a learning rate decaying from $5\times 10^{-5}$ to $1 \times 10^{-6}$.

 \begin{figure*}
    \centering
    \includegraphics[width=0.95\textwidth]{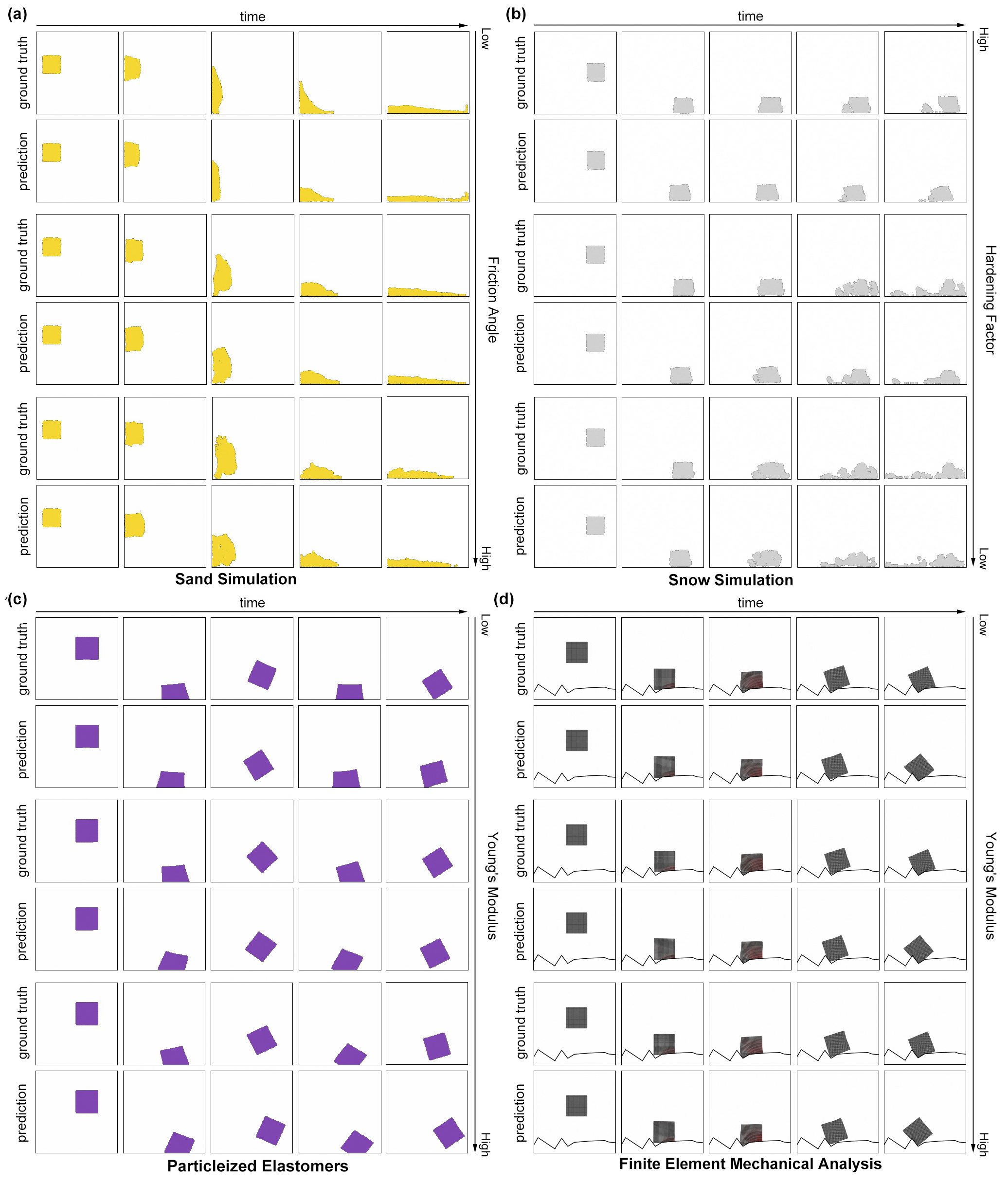}
    \caption{The dynamics of \textsc{Sand}, \textsc{Snow}, \textsc{Elastomers} and \textsc{FEM} predicted by GPE. (a) The motion of a sand pile with low, medium and high friction angles. (b) Deformation of snow masses with different hardening coefficients after collision and compression. (c) Drop deformation of elastomers with different hardness. (d) FEM results of collision deformation of an elastomer with different Young's modulus on an irregular obstacle, with the internal stress distribution shown in red.}
    \label{fig:my_fig}
\end{figure*}

\textbf{Boundary nodes.} GPE discretizes all boundaries or obstacles into a series of dense scatter nodes called ``boundary nodes''. Collision dynamics need to be learned through message passing between the material nodes and the boundary nodes. Handling the boundary conditions in this way allows the learned model to generalize to scenarios with more complex boundaries and obstacles, which will be demonstrated later. 

\subsection{Results}

\begin{figure}[h]
    \centering
    \includegraphics[width=0.35\textwidth]{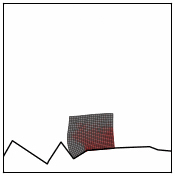}
    \caption{The stress distribution during the deformation of the meshed elastic body is shown in red.} 
    \label{fig:stress}
\end{figure}

In Figure~\ref{fig:my_fig}, we visualize the \textsc{Sand}, \textsc{Snow}, \textsc{Elastomers} and \textsc{FEM} dynamics predicted by GPE, where each row of a subfigure depicts the dynamics of one physical system along time. We observe that the long-range roll-out trajectories predicted by GPE are very similar to the ground truths, even if the models are trained only by single-step error. For the \textsc{FEM} domain, the distribution of internal stress is shown in red and enlarged in Figure~\ref{fig:stress}. It is worth noticing that these domains are simulated by a unified model framework, where the graph representation depends on how the system is discretized.

Moreover, we demonstrate the ability of GPE to simulate physical systems with unseen parameters. In each subfigure, the top two rows with lowest physical parameters and the bottom two rows with highest physical parameters are seen in the training set, whereas the middle two rows with intermediate physical parameters are unseen. We observe that GPE is able to simulate physical systems with unseen parameters.

\subsubsection{Comparison with baselines}
To quantify the performance of the simulation on unseen materials, we compare GPE with three baselines, DPI-Net~\cite{li2018learning} with hierarchy, GNS~\cite{sanchez2020learning} and ConvNet~\cite{ummenhofer2019lagrangian}. We calculate the average MSE of each time step in roll-out for the aforementioned domains on unseen physical parameters only. For GPE and DPI-Net, nearest neighbor graph are used for the \textsc{Sand}, \textsc{Snow} and \textsc{Fluid} domains, while multi-scale grid graphs are used for the \textsc{Elatomer} and \textsc{FEM} domains. For GNS and ConvNet, the graph topologies for all domains are consistent with original papers. 

\begin{table}[]
    \centering
    \begin{tabular}{c c c c c}
    \hline
         Methods & DPI-Net & GNS & ConvNet & \textbf{GPE} \\
    \hline
        \textsc{Sand} & 5.15 & 3.80 & 3.51 & \textbf{2.37} \\
        \textsc{Snow} & 4.91 & 5.49 & 5.24 & \textbf{3.94} \\
        \textsc{Elastomer} & 2.07 & 3.58 & 2.82 & \textbf{1.35} \\
        \textsc{FEM} & 1.76 & 5.03 & 4.96 & \textbf{1.27} \\
        \textsc{Fluid} & 3.31 & 3.78 & \textbf{2.75} & 3.22 \\
    \hline
    \end{tabular}
    \caption{The average MSE ($\times 10^{-3}$) between the ground truth and rollout trajectories.}
    \setlength{\belowcaptionskip}{-0.cm}
    \label{tab:my_table}
\end{table}

As shown in Table~\ref{tab:my_table}, GPE outperforms the baseline methods in most domains, demonstrating its better ability to generalize to unseen systems. 

\subsubsection{Effects of momentum conservation}
\begin{figure}[h]
    \centering
    \includegraphics[width=0.38\textwidth]{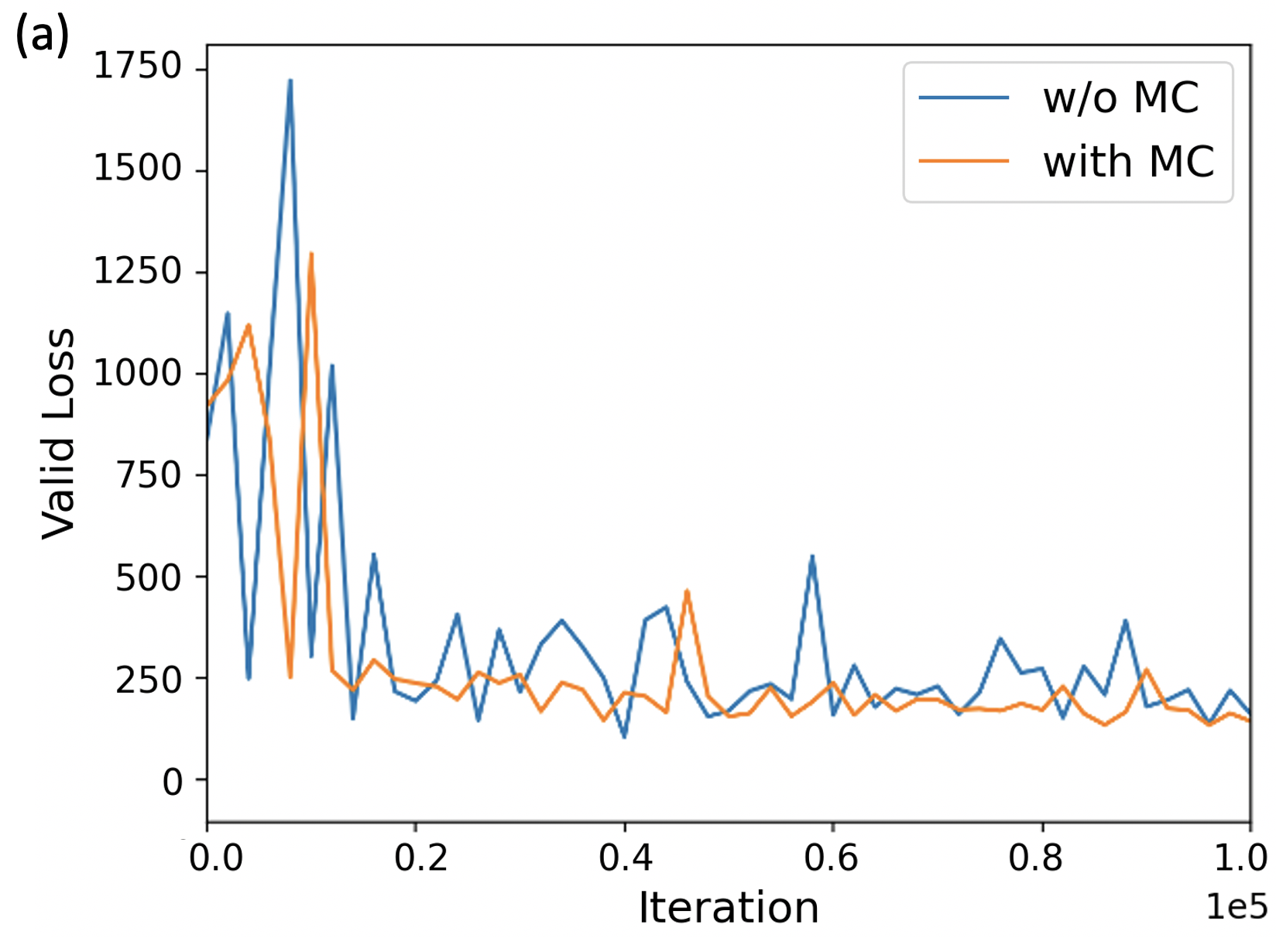}
    \includegraphics[width=0.38\textwidth]{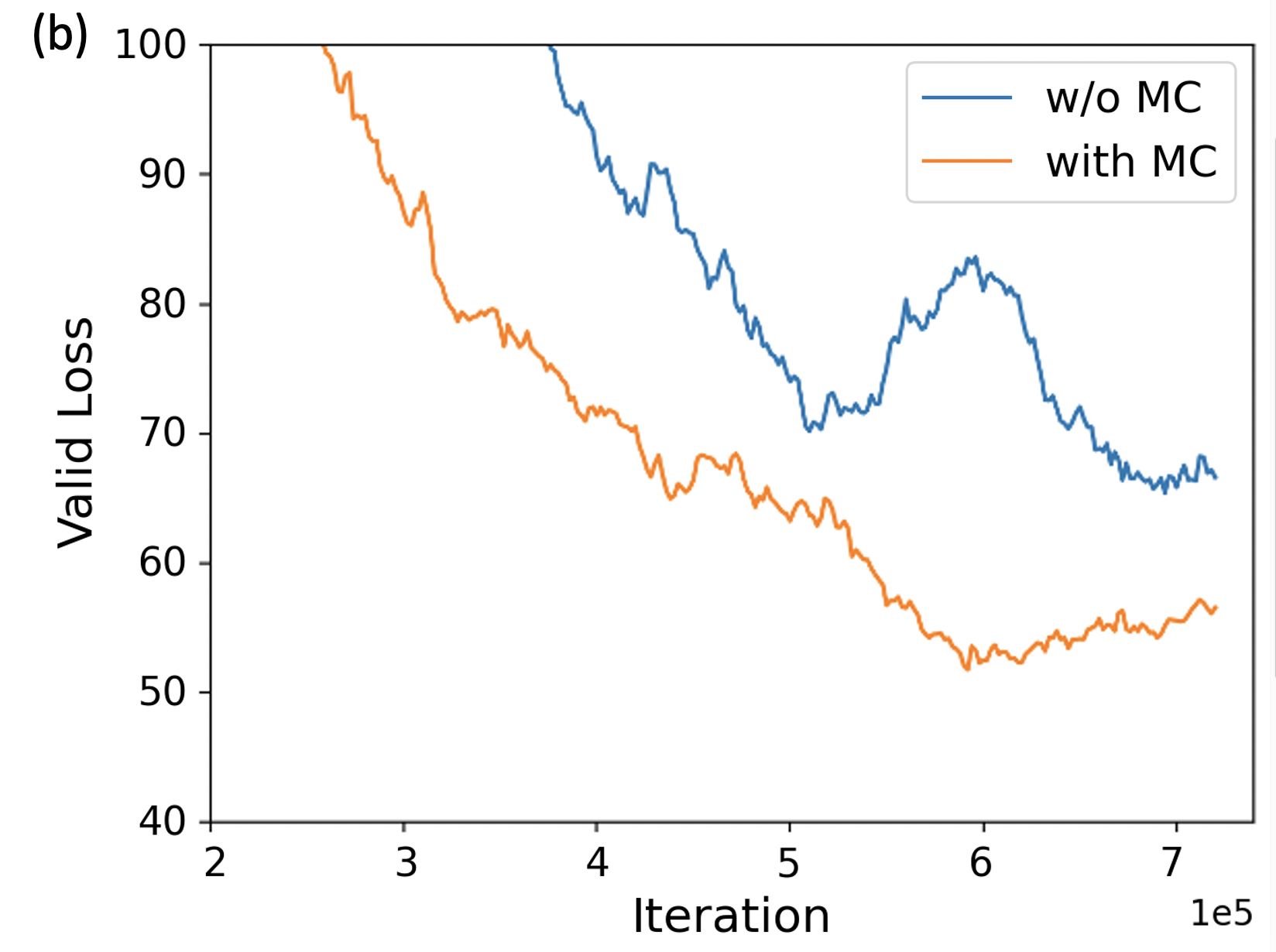}
    \caption{Validation losses of GPE with and without momentum conservation. (a): First 100,000 training iterations. (b): 200,000-700,000 training iterations.} 
    \label{fig:momentum_conservation}
\end{figure}

We demonstrate the effect of momentum conservation on \textsc{Fluid} domain. 
By introducing momentum conservation, only half of the messages on the edges are computed, leading to a 37\% reduction of forward computation time. 
As shown in Figure~\ref{fig:momentum_conservation}.(a), GPE with momentum conservation is more stable at the first 100,000 training steps, compared to GPE without such a constraint. Eventually, GPE with momentum conservation converge to a model with lower validation loss, as shown in Figure~\ref{fig:momentum_conservation}.(b).

\subsubsection{Generalizability of GPE}
\begin{figure*}
    \centering
    \includegraphics[width=0.75\textwidth]{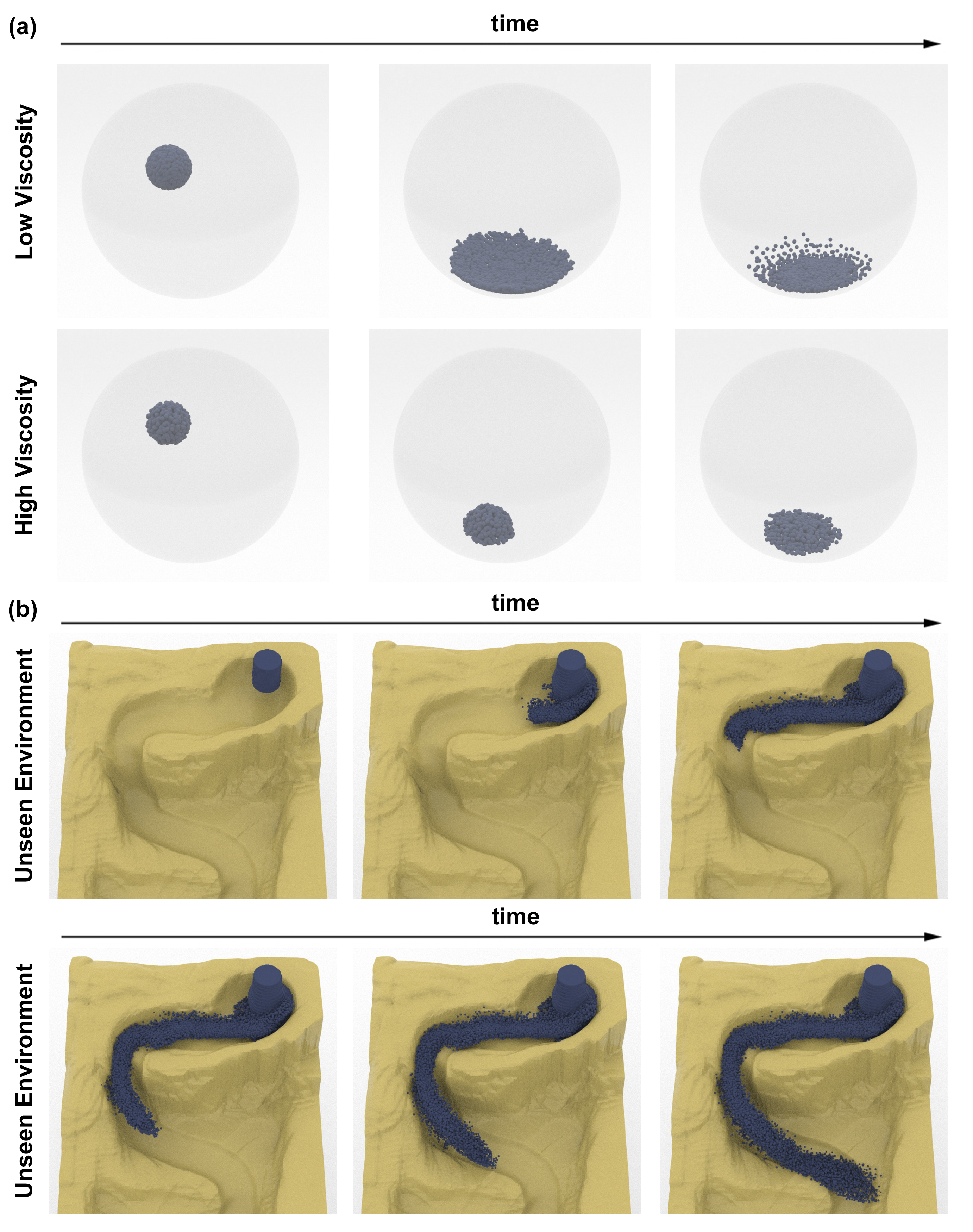}
    \caption{The dynamics of \textsc{Fluid} predicted by GPE. (a) The training set contains the falling of viscous liquids in a 2D spherical boundary. (b) The learned GPE can simulate the flow of liquids in a complex 3D canyon with a much larger scale.}
    \label{fig:canyon}
\end{figure*}

We test whether GPE is able to generalize to larger and more complex scenarios than the training data. As shown in Figure~\ref{fig:canyon}.(a), the training set of the \textsc{Fluid} domain contains the trajectories of liquids with different viscosities falling then colliding with a 2D spherical boundary. Only 4000 liquid nodes and 3000 boundary nodes are included in the system. We test the trained GPE on a much more difficult scenario --- falling of liquids along a complex canyon in three dimensional space. The number of liquid nodes and boundary nodes are 150,000 and 50,000 respectively, which is much larger than the training set. We found that GPE trained by 2D spherical boundaries can be used to accurately simulate the flow of liquids in 3D curved canyons as well as waterfalls, which proves that our model learns the essential law of dynamics of physical systems, rather than remembering training data.

%% file: conclusion.tex
In this paper, we proposed GPE, a graph-based learnable physical engine that simulates the dynamics of complex systems by learning the message passing between nodes. Our experiments show that GPE can simulate complex evolutionary trajectories in a variety of domains,
and has good generalization ability to unseen materials. 
We also introduce momentum conservation to reduce computational cost
and stabilize the training process.
Thanks to the universality, expressiveness and generalizability, graph-based learnable physics engines show great potential for applications in high precision simulation of complex systems and inverse problem solving.

\textbf{Limitations and Future Work.} There is still some work to be explored in using machine learning to solve major physical problems. 
First, compared to traditional physical engines, the computational speed and memory consumption of GNN-based physical engines are still at a disadvantage. These disadvantages will highly limit the applicability and scalability of GNN-based engines. It would be of great importance to use the powerful expressiveness and computational features of neural networks to accelerate the computational process and increase the scale of simulation. 
Second, the lack of high-quality real-world training data remains a major obstacle limiting the utility of learnable physics engines. Incorporating fundamental physic laws and first principles into the design or training of neural networks might help to reduce data consumption and improve training efficiency.
